\documentclass{article}

\usepackage[preprint]{neurips_2023}
\usepackage[utf8]{inputenc}
\usepackage[T1]{fontenc}
\usepackage{url}
\usepackage{booktabs}
\usepackage{amsfonts}
\usepackage{amsmath}
\usepackage{amssymb}
\usepackage{nicefrac}
\usepackage{microtype}
\usepackage{xcolor}
\usepackage{multirow}
\usepackage[colorlinks=true,
            linkcolor=blue,
            citecolor=blue,
            urlcolor=blue]{hyperref}
\usepackage{graphicx}
\usepackage{xspace}
\usepackage{enumitem}
\usepackage{tikz}
\usetikzlibrary{arrows.meta,positioning,decorations.pathreplacing}
\definecolor{h2oyellow}{HTML}{F5C518}
\usepackage{pgfplots}
\pgfplotsset{compat=1.18}
\usepackage{listings}

\newcommand{\tabh}{\mbox{\textit{TabH2O}}\xspace}
\newcommand{\tabhv}{\mbox{\textit{TabH2O v1}}\xspace}
\newcommand{\tabhvone}{\mbox{\textit{TabH2O v1.1}}\xspace}

\title{TabH2O: A Unified Foundation Model for Tabular Prediction}
\author{
  Pascal Pfeiffer
  $\quad$Dmitry Gordeev
  $\quad$Mathias M\"uller
  $\quad$Laura Fink
  $\quad$Joan Salvà Soler\\
  \textbf{Mark Landry}
  $\quad$\textbf{Branden Murray}
  $\quad$\textbf{Marcos V. Conde}
  $\quad$\textbf{Sri Satish Ambati}\\
  H2O.ai\\
  \texttt{\{firstname.lastname, sri\}@h2o.ai}\\
  \href{https://tabh2o.h2oai.com}{\tikz[baseline=(tabh2ourl.base)]\node[fill=yellow!85!orange, rounded corners=3pt, inner xsep=5pt, inner ysep=1.8pt, text=black] (tabh2ourl) {\texttt{https://tabh2o.h2oai.com}};}
}

\begin{document}

\maketitle

\section{Abstract}

We present \tabh, a foundation model for tabular data that performs classification and regression in a single forward pass via in-context learning. \tabh builds on the TabICL architecture~\cite{qu2025tabicl, qu2026tabiclv2} with several key modifications: (1)~\emph{unified training}---a single model handles both classification and regression via a dual-head architecture, eliminating the need for separate models and reducing total pretraining cost; (2)~\emph{single-stage pretraining}---training stability improvements (bounded scalable softmax, inter-stage normalization, learnable residual scaling, logit soft-capping) eliminate the need for multi-stage curriculum learning, enabling training with full-length sequences from the start; and (3)~\emph{noise-aware pretraining}---synthetic datasets include explicit noise dimensions to teach the model robustness to irrelevant features.

We evaluate \tabhvone (29.2M parameters) on the TALENT benchmark~\cite{liu2024talent} (300 datasets), where it achieves an average rank of 2.54 out of 6 evaluated methods, outperforming tuned CatBoost (4.00), H2O AutoML (4.28), and LightGBM (4.98), competitive with TabPFN~v2.6 (2.80), and behind TabICL~v2 (2.13), while placing in the top~3 on 80\% of datasets. On TabArena~\cite{erickson2025tabarena} (51 datasets) it reaches average rank 1.94 among three foundation models (ahead of TabPFN~v2.6 and TabICL~v2; Section~\ref{sec:tabh2o_v11}).

\paragraph{Intended scope.} \tabh is well-suited to tabular datasets up to roughly $500{,}000$ rows and $100$ features, to workflows that prefer a single model serving many use cases rather than loading task-specific models in production, and to batch inference. Beyond saving time on baseline benchmarking for model performance, imputation, and anomaly detection, early use cases include applied tabular workloads in cybersecurity, fraud prevention, and customer experience. It is, today, less well-suited to single-sample real-time inference, to ultra-large datasets that materially exceed the training envelope, to time-series tasks (currently supported experimentally, with first-class native handling planned for the next iteration), and to multi-modal data such as text, images, or audio.

\section{Introduction}

Tabular data powers critical decisions in business---credit scoring, demand forecasting, clinical trials, fraud detection. Yet building a good model for a new dataset still requires significant effort: feature engineering, model selection, hyperparameter tuning, and weeks of iteration.

Foundation models transformed NLP and computer vision by pretraining a single model on broad data and transferring it to downstream tasks~\cite{devlin2018bert, brown2020language, dosovitskiy2020vit}. Recent work has brought this paradigm to tabular data. TabPFN~\cite{hollmann2023tabpfn} demonstrated that a transformer pretrained on synthetic datasets can perform in-context learning (ICL) for tabular classification. TabPFN~v2 ~ \cite{hollmann2025tabpfnv2} extended this to regression and larger datasets and was published in Nature. TabICL~\cite{qu2025tabicl} introduced a scalable column-then-row architecture enabling ICL on datasets with up to 500K samples, and TabICL~v2~\cite{qu2026tabiclv2} further improved performance through architectural innovations including query-aware scalable softmax (QASSMax), a richer synthetic data prior, and the Muon optimizer~\cite{jordan2024muon}.

\tabh builds directly on this lineage---specifically on the TabICL~v2 architecture and training methodology. We gratefully acknowledge the foundational contributions of TabPFN, TabPFN~v2, TabICL, and TabICL~v2 to this model family. Our work focuses on engineering improvements that make the model more practical for deployment:

\begin{itemize}[nosep]
    \item \textbf{Unified classification and regression} in a single model, reducing total pretraining cost.
    \item \textbf{Single-stage pretraining} enabled by training stability mechanisms that eliminate multi-stage curriculum learning---the model trains with full-length sequences from step one (up to 12{,}288 rows for \tabhv and 14{,}336 for \tabhvone).
    \item \textbf{Noise-aware pretraining} with explicit uninformative feature dimensions.
    \item \textbf{A production-ready API} with memory-efficient inference for datasets up to 500K rows.
\end{itemize}

\tabh was developed using the publicly available TabICL~v1 codebase as a starting point, with independent implementations of our modifications. In this report, we detail our architecture, pretraining procedure, key differences from prior work, and benchmark results.

\section{Related Work}
\label{sec:related}

\paragraph{Gradient-boosted decision trees.}
GBDTs---XGBoost~\cite{chen2016xgboost}, LightGBM~\cite{ke2017lightgbm}, CatBoost~\cite{prokhorenkova2018catboost}---have dominated tabular prediction for over a decade. They require per-dataset hyperparameter tuning and produce per-task models with no cross-task transfer.

\paragraph{AutoML systems.}
H2O AutoML~\cite{h2o2024automl} and similar systems (e.g., AutoGluon~\cite{erickson2020autogluon}) automate model selection and ensembling but demand substantial compute budgets for each new dataset.

\paragraph{Tabular foundation models.}
TabPFN~\cite{hollmann2023tabpfn} introduced Prior-Data Fitted Networks for tabular ICL, pretrained on synthetic data from structural causal models~\cite{pearl2009causality}. TabPFN~v2~\cite{hollmann2025tabpfnv2} scaled to 10K samples with alternating row--column attention at $O(n^2 m + nm^2)$ complexity. TabICL~\cite{qu2025tabicl} reduced this to $O(n^2 + nm^2)$ via a two-stage column-then-row architecture and introduced a Set Transformer~\cite{lee2019isab}-based column embedding. TabICL~v2~\cite{qu2026tabiclv2} achieved state-of-the-art results through QASSMax, the Muon optimizer~\cite{jordan2024muon}, target-aware embedding, repeated feature grouping, a richer synthetic data prior with 8 function types, and an ExtraTrees~\cite{geurts2006extratrees}-based quality filter.

\paragraph{Tabular benchmarks.}
TALENT~\cite{liu2024talent} provides standardized evaluation across 300 tabular datasets. TabArena~\cite{erickson2025tabarena} offers a complementary 51-dataset leaderboard (38 classification with $\leq10$ classes, 13 regression). We use TALENT as our primary benchmark due to its larger scale and diversity.

\section{Architecture}
\label{sec:architecture}

\begin{figure}[p]
  \centering
  \begin{tikzpicture}[
    fcell/.style={draw=gray!50, fill=gray!10, minimum width=0.36cm, minimum height=0.32cm, inner sep=0pt, line width=0.3pt},
    hcell/.style={fcell, fill=gray!22, font=\sffamily\fontsize{6.5}{7.5}\selectfont\bfseries, text=gray!85!black},
    ycell/.style={fcell, fill=h2oyellow, font=\sffamily\fontsize{7}{8}\selectfont\bfseries, text=black},
    qcell/.style={fcell, fill=gray!20, font=\sffamily\fontsize{8.5}{9}\selectfont\bfseries, text=gray!50!black},
    tensor/.style={draw=gray!50, fill=gray!10, line width=0.45pt, rounded corners=2pt,
                   minimum width=4.8cm, minimum height=0.65cm, align=center,
                   font=\sffamily\fontsize{8.5}{10}\selectfont, text=gray!80!black},
    stagebox/.style={draw=gray!35, fill=h2oyellow, line width=0.9pt, rounded corners=8pt,
                     minimum width=13.0cm, minimum height=3.60cm, inner sep=0pt},
    attnbox/.style={draw=gray!40, fill=white, line width=0.4pt, rounded corners=2pt,
                    minimum width=3.4cm, minimum height=2.80cm, inner sep=2pt},
    mcell/.style={draw=gray!55, fill=gray!10, minimum width=0.20cm, minimum height=0.18cm,
                  inner sep=0pt, line width=0.3pt},
    mhicell/.style={draw=h2oyellow!60!orange, fill=h2oyellow!25, minimum width=0.20cm,
                    minimum height=0.18cm, inner sep=0pt, line width=0.45pt},
    rowvec/.style={draw=gray!55, fill=gray!10, minimum width=0.85cm, minimum height=0.20cm,
                   inner sep=0pt, line width=0.3pt},
    rowvecY/.style={draw=h2oyellow!60!orange, fill=h2oyellow!85, minimum width=0.20cm,
                    minimum height=0.20cm, inner sep=0pt, line width=0.4pt,
                    font=\sffamily\fontsize{6}{7}\selectfont\bfseries, text=black},
    rowvecQ/.style={draw=gray!60, fill=gray!22, minimum width=0.20cm, minimum height=0.20cm,
                    inner sep=0pt, line width=0.4pt,
                    font=\sffamily\fontsize{7}{8}\selectfont\bfseries, text=gray!55!black},
    flow/.style={-{Stealth[length=7pt, width=6pt]}, line width=0.9pt, gray!55},
    attnarr/.style={-{Stealth[length=2.8pt, width=2.4pt]}, line width=0.5pt,
                    draw=h2oyellow!40!black},
    iclarr/.style={-{Stealth[length=3pt, width=2.6pt]}, line width=0.45pt,
                   draw=h2oyellow!40!black},
    title/.style={font=\sffamily\fontsize{8.5}{10}\selectfont\bfseries, text=black, align=left},
    desc/.style={font=\sffamily\fontsize{8.5}{10.5}\selectfont, text=black, align=left},
    side/.style={font=\sffamily\fontsize{8}{9.5}\selectfont, text=gray!72!black, align=left},
    sidebold/.style={font=\sffamily\fontsize{8}{9.5}\selectfont\bfseries, text=gray!85!black, align=left},
    insetlbl/.style={font=\sffamily\fontsize{6.5}{7.5}\selectfont\bfseries, text=gray!75!black, align=center},
    insetnote/.style={font=\sffamily\fontsize{6.5}{7.5}\selectfont, text=gray!75!black, align=center},
    bigtitle/.style={font=\sffamily\fontsize{9}{10.5}\selectfont\bfseries, text=gray!85!black, align=center},
  ]

  \node[bigtitle, anchor=south] at (0, 0.30) {Input dataset $X \in \mathbb{R}^{n \times m}$};

  \node[hcell] at (-0.72, 0)   {$f_1$};
  \node[hcell] at (-0.36, 0)   {$f_2$};
  \node[hcell] at ( 0.00, 0)   {$f_3$};
  \node[hcell] at ( 0.36, 0)   {$f_4$};
  \node[hcell, fill=h2oyellow!50] at (0.72, 0) {$y$};

  \foreach \r/\lab in {1/A, 2/B, 3/A} {
    \pgfmathsetmacro{\yy}{-0.32*\r}
    \foreach \xc in {-0.72, -0.36, 0.00, 0.36} { \node[fcell] at (\xc, \yy) {}; }
    \node[ycell] at (0.72, \yy) {\lab};
  }
  \foreach \r in {4, 5} {
    \pgfmathsetmacro{\yy}{-0.32*\r-0.10}
    \foreach \xc in {-0.72, -0.36, 0.00, 0.36} { \node[fcell] at (\xc, \yy) {}; }
    \node[qcell] at (0.72, \yy) {?};
  }

  \node[side, anchor=west] at (1.20, -0.64) {\textit{labeled training rows}\;($y_i$ given)};
  \node[side, anchor=west] at (1.20, -1.62) {\textit{query rows}\;($y_i$ predicted)};

  \draw[flow] (0, -1.95) -- (0, -3.30);
  \node[side, anchor=west] at (1.20, -2.10) {\textbf{repeated feature grouping}};
  \node[side, anchor=west] at (1.20, -2.42) {\quad{}circular shifts $\to$ groups of 3};
  \node[side, anchor=west] at (1.20, -2.74) {\textbf{linear projection}\;$\mathbb{R}^3 \to \mathbb{R}^d$};
  \node[side, anchor=west] at (1.20, -3.06) {\textbf{$+$ target-aware embedding}\;(training rows)};

  \node[tensor] (e2) at (0, -3.65) {$E \in \mathbb{R}^{n \times m \times d}$\quad{}per-sample, per-feature embeddings};
  \draw[flow] (0, -4.00) -- (0, -4.50);

  \node[stagebox] (s1) at (0, -6.30) {};
  \node[title, anchor=north west] at ([shift={(8pt,-8pt)}]s1.north west)
    {\textsc{Stage 1}\quad TF$_{\text{col}}$\;---\;column-wise Set Transformer (ISAB)};
  \node[desc, anchor=north west, text width=8.5cm] at ([shift={(8pt,-26pt)}]s1.north west)
    {Each feature column attends \textbf{within itself} across rows. ISAB inducing points are summarized from \textbf{train rows only}; all rows (train\,+\,query) then read back from those points. Cost $\mathcal{O}(n\,k\,m)$, output shape $\mathbb{R}^{n \times m \times d}$.};

  \node[attnbox] (s1in) at ([xshift=-0.22cm,yshift=0]s1.east) [anchor=east] {};
  \node[insetlbl, anchor=north] at ([yshift=-2pt]s1in.north) {column-wise (ISAB)};
  \begin{scope}[shift={(s1in.center)}, yshift=0]
    \foreach \my/\sty in {0.45/h2oyellow, 0.27/h2oyellow, 0.09/h2oyellow,
                          -0.15/gray, -0.33/gray} {
      \node[draw=gray!55, fill=\sty!50, minimum width=0.20cm, minimum height=0.16cm,
            inner sep=0pt, line width=0.3pt] at (-1.00, \my) {};
    }
    \foreach \my in {0.31, 0.07} {
      \node[circle, draw=h2oyellow!60!orange, fill=h2oyellow, inner sep=0pt,
            minimum size=0.14cm, line width=0.4pt] at (-0.20, \my) {};
    }
    \foreach \my in {0.45, 0.27, 0.09} {
      \draw[attnarr, gray!60] (-0.87, \my) to[bend left=10] (-0.30, 0.19);
    }
    \foreach \my/\sty in {0.45/h2oyellow, 0.27/h2oyellow, 0.09/h2oyellow,
                          -0.15/gray, -0.33/gray} {
      \node[draw=gray!55, fill=\sty!50, minimum width=0.20cm, minimum height=0.16cm,
            inner sep=0pt, line width=0.3pt] at (1.00, \my) {};
    }
    \foreach \my in {0.45, 0.27, 0.09, -0.15, -0.33} {
      \draw[attnarr] (-0.10, 0.19) to[bend left=10] (0.87, \my);
    }
    \node[insetnote, anchor=south, text width=3.2cm, align=center] at (0, -1.20)
      {inducing pts summarize\\\textsc{train} rows; all read back};
  \end{scope}

  \draw[flow] (0, -8.15) -- (0, -8.65);

  \node[stagebox] (s2) at (0, -10.50) {};
  \node[title, anchor=north west] at ([shift={(8pt,-8pt)}]s2.north west)
    {\textsc{Stage 2}\quad TF$_{\text{row}}$\;---\;row-wise transformer (RoPE\,+\,CLS)};
  \node[desc, anchor=north west, text width=8.5cm] at ([shift={(8pt,-26pt)}]s2.north west)
    {Each row attends \textbf{across its own features} with rotary positional embeddings (RoPE). Applied uniformly to every row\,---\,no train/query distinction. Learnable CLS tokens then aggregate features into a single row vector $h_i \in \mathbb{R}^d$.};

  \node[attnbox] (s2in) at ([xshift=-0.22cm,yshift=0]s2.east) [anchor=east] {};
  \node[insetlbl, anchor=north] at ([yshift=-2pt]s2in.north) {row-wise};
  \begin{scope}[shift={(s2in.center)}, yshift=4pt]
    \foreach \mx in {-0.30, -0.10, 0.10, 0.30} {
      \foreach \my in {0.20, 0.0, -0.20} {
        \node[mcell] at (\mx, \my) {};
      }
    }
    \foreach \mx in {-0.30, -0.10, 0.10, 0.30} { \node[mhicell] at (\mx, 0) {}; }
    \draw[attnarr] (-0.20, 0) -- (-0.14, 0);
    \draw[attnarr] (0.14, 0) -- (0.20, 0);
    \draw[attnarr, bend left=55] (-0.30, 0) to (0.30, 0);
    \node[insetnote, anchor=south, text width=3.2cm, align=center] at (0, -1.20)
      {attention within each row\\(applied to all rows)};
  \end{scope}

  \draw[flow] (0, -12.30) -- (0, -12.75);
  \node[tensor, minimum width=4.0cm] at (0, -13.10) {$h \in \mathbb{R}^{n \times d}$\quad row representations};
  \draw[flow] (0, -13.45) -- (0, -13.90);

  \node[stagebox] (s3) at (0, -15.70) {};
  \node[title, anchor=north west] at ([shift={(8pt,-8pt)}]s3.north west)
    {\textsc{Stage 3}\quad TF$_{\text{icl}}$\;---\;dataset-wise in-context learning};
  \node[desc, anchor=north west, text width=8.5cm] at ([shift={(8pt,-26pt)}]s3.north west)
    {Block-masked attention \textbf{across rows}: K/V is restricted to \textbf{train rows} at every block. So train rows attend among themselves, and each query attends to all train rows\,---\,but never to other queries. The labeled context conditions every prediction in a single forward pass.};

  \node[attnbox] (s3in) at ([xshift=-0.22cm,yshift=0]s3.east) [anchor=east] {};
  \node[insetlbl, anchor=north] at ([yshift=-2pt]s3in.north) {across rows (K/V = train)};
  \begin{scope}[shift={(s3in.center)}, yshift=0]
    \foreach \my/\lab in {0.55/A, 0.37/B, 0.19/A} {
      \node[rowvec, minimum width=0.65cm] at (-0.65, \my) {};
      \node[rowvecY] at (-0.20, \my) {\lab};
    }
    \foreach \my in {-0.07, -0.27} {
      \node[rowvec, minimum width=0.65cm] at (0.35, \my) {};
      \node[rowvecQ] at (0.80, \my) {?};
    }
    \node[font=\fontsize{11}{12}\selectfont, text=gray!70!black] at (-1.20, 0.37) {$\circlearrowleft$};
    \draw[iclarr] (-0.05, 0.37) to[out=-30, in=170] (0.55, -0.07);
    \draw[iclarr] (-0.05, 0.37) to[out=-50, in=170] (0.55, -0.27);
    \node[insetnote, anchor=south, text width=3.2cm, align=center] at (0, -1.20)
      {train\,$\leftrightarrow$\,train; queries\,$\to$\,train only};
  \end{scope}

  \draw[flow] (0, -17.55) -- (0, -18.05);

  \node[bigtitle, anchor=north] at (0, -18.15) {Predictions $\hat{y}$ for query rows};
  \node[side, anchor=north, align=center] at (0, -18.45)
    {classification head\;($\to$ logits)\quad\textbar\quad regression head\;($\to$ 999 quantiles)};

  \end{tikzpicture}
  \caption{\textbf{TabH2O architecture.} The model reads the entire dataset (training rows with labels and unlabeled query rows) in a single forward pass. Preprocessing forms per-cell embeddings via repeated feature grouping, a linear projection, and target-aware embedding for training rows. Three transformer stages then apply attention along three orthogonal axes: \textsc{Stage~1} attends \emph{within each column}\,---\,inducing points are summarized from \emph{train rows only}, and all rows read back from them; \textsc{Stage~2} attends \emph{within each row} symmetrically (no train/query distinction), aggregating to a single vector $h_i$ per row; \textsc{Stage~3} applies block-masked attention \emph{across rows} where keys and values are restricted to train rows, so train rows attend among themselves and each query attends to all train rows but never to another query. Architecture follows TabICL\,v2~\cite{qu2026tabiclv2}, with modifications for unified classification/regression and training stability (Section~\ref{sec:differences}).}
  \label{fig:architecture}
\end{figure}
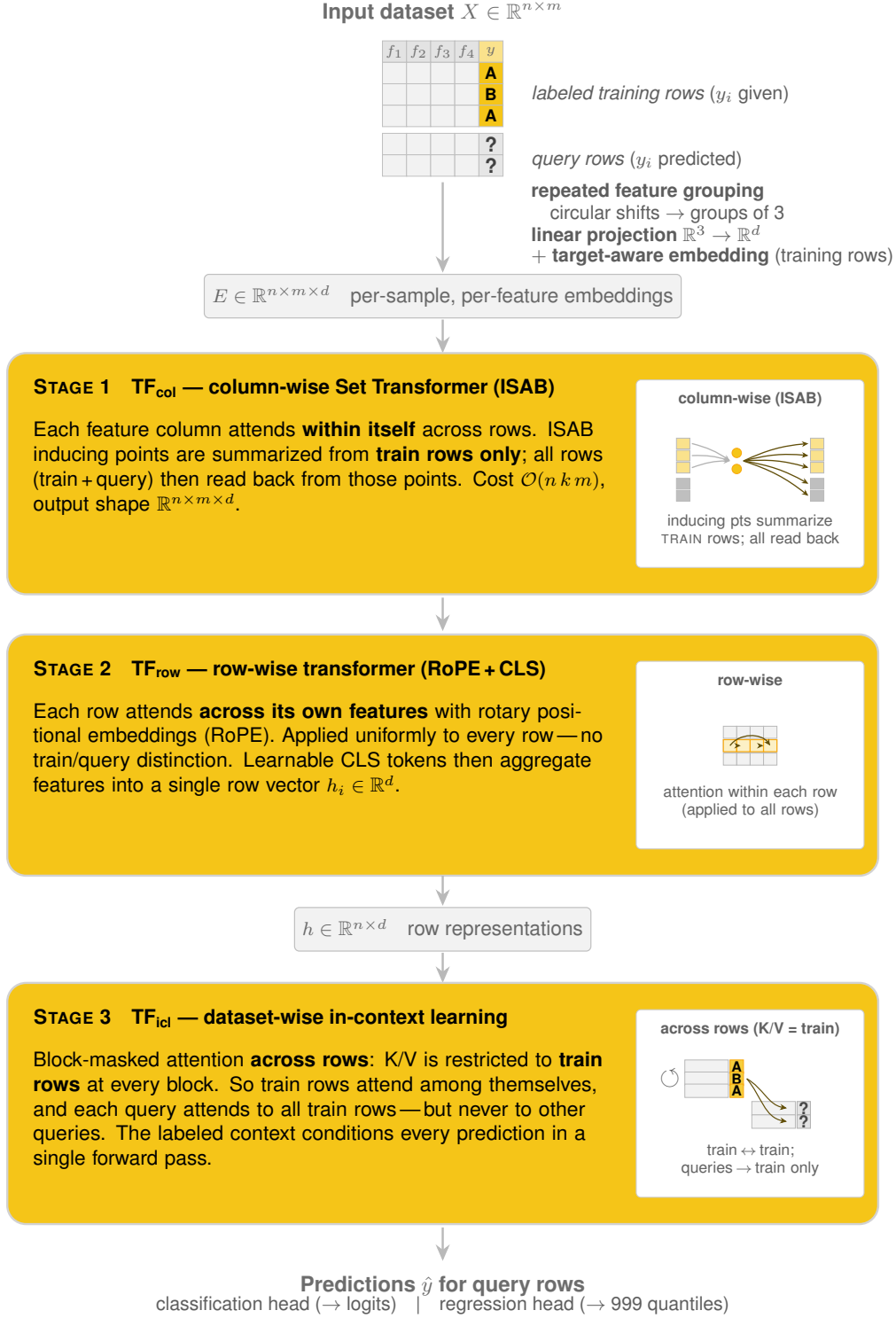

\tabh follows the three-stage transformer architecture introduced by TabICL~\cite{qu2025tabicl}: column embedding, row interaction, and dataset-wise in-context learning (Figure~\ref{fig:architecture}). Each stage processes tabular data at increasing granularity---from individual features, to rows, to the entire dataset.

\paragraph{Stage 1: Column Embedding.} A shared Set Transformer~\cite{lee2019isab} processes each column independently using induced self-attention blocks (ISAB). Within an ISAB, $k$ learnable inducing vectors first attend to the \emph{train rows only} of that column, summarizing them; all rows (train and query alike) then attend to those inducing points. This asymmetry keeps query rows from contaminating the column summary while still letting them read it. Features are grouped into sets of 3 via repeated circular shifts following TabICL~v2~\cite{qu2026tabiclv2}, and target-aware embeddings inject label information into train-row representations from the start. The complexity is $O(nkm)$.

\paragraph{Stage 2: Row Interaction.} Cross-feature attention with rotary position embeddings (RoPE)~\cite{su2024rope} captures interactions between features within each row. Learnable CLS tokens aggregate feature information into a fixed-dimensional row representation. The complexity is $O(nm^2)$.

\paragraph{Stage 3: In-Context Learning.} A standard self-attention transformer~\cite{vaswani2017attention} operates across rows with a block mask: at every layer, keys and values are restricted to the train rows. As a consequence, train rows attend to one another bidirectionally (full attention within the train block), and each query row attends to all train rows---never to other queries or to itself. This is what enables in-context learning: the labeled context conditions every prediction in a single forward pass. A classification head and regression head produce task-specific outputs. The complexity is $O(n^2)$.

The overall complexity is $O(n^2 + nm^2)$, identical to TabICL. Key parameters are summarized in Table~\ref{table:architecture_params}.

\begin{table}[h]
    \centering
    \caption{\textbf{Key model parameters.}}
    \label{table:architecture_params}
    \begin{tabular}{lc}
        \toprule
        \textbf{Parameter} & \textbf{Value} \\
        \midrule
        Total parameters & 29.2M \\
        Embedding dimension ($d$) & 128 \\
        ICL dimension ($d \times$ CLS tokens) & 512 \\
        Column embedding blocks & 3 (ISAB, 8 heads, 128 inducing vectors) \\
        Row interaction blocks & 3 (8 heads, RoPE base 100{,}000) \\
        ICL blocks & 12 (4 heads) \\
        Feed-forward expansion & $2\times$ \\
        Normalization & RMSNorm~\cite{zhang2019rmsnorm}, pre-norm \\
        \bottomrule
    \end{tabular}
\end{table}

\section{Key Differences from TabICL v2}
\label{sec:differences}

While \tabh shares the core three-stage architecture with TabICL~v2~\cite{qu2026tabiclv2}, we introduce several modifications aimed at simplifying the training recipe, improving stability, and making deployment more efficient. Table~\ref{tab:differences} summarizes the differences.

\begin{table}[h]
    \centering
    \caption{\textbf{Key differences between TabH2O and TabICL v2.}}
    \label{tab:differences}
    \small
    \begin{tabular}{lcc}
        \toprule
        \textbf{Feature} & \textbf{TabICL v2} & \textbf{TabH2O} \\
        \midrule
        \multicolumn{3}{l}{\textit{Training}} \\
        Classification/regression & Separate models & \textbf{Unified (single model)} \\
        Curriculum stages & 3 (500K + 40K + 10K steps) & \textbf{1 (both versions)} \\
        Optimizer updates & 550K & \textbf{100K (v1); 50K (v1.1)} \\
        Max seq.\ length (stage 1) & 1{,}024 & \textbf{12{,}288 (v1); 14{,}336 (v1.1)} \\
        Total pretraining datasets & $\sim$35M & \textbf{$\sim$6.4M (v1); $\sim$12.8M (v1.1)} \\
        Optimizer & Muon & Muon \\
        Weight decay & 0.01 (cautious) & 0.05 (standard) \\
        \midrule
        \multicolumn{3}{l}{\textit{Architecture (stability)}} \\
        Scalable softmax & QASSMax & \textbf{Bounded QASSMax} \\
        Inter-stage normalization & --- & \textbf{RMSNorm} \\
        Residual scaling & Standard & \textbf{Learnable $\lambda$ + $x_0$ blending} \\
        Logit soft-capping & --- & \textbf{$\tanh$-based ($c{=}30$)} \\

        \midrule
        \multicolumn{3}{l}{\textit{Synthetic Data}} \\
        Function types & 8 & 8 (same) \\
        Noise dimensions & --- & \textbf{Up to 3 (v1); 2 (v1.1) per node} \\
        Max ``easy'' AUC filter & --- & \textbf{$\leq 0.99$} \\
        \bottomrule
    \end{tabular}
\end{table}

\subsection{Unified Classification and Regression}
\label{sec:unified}

TabICL~v2 trains separate models for classification and regression, as do TabPFN~v2 and most prior tabular foundation models. \tabh instead uses a \textbf{single unified model} with dual output heads: a classification head (cross-entropy loss over class logits) and a regression head (pinball loss over 999 quantile predictions). During training, each mini-batch contains a mix of classification and regression datasets (80\%/20\% split via \texttt{reg\_prob=0.2}), and the appropriate head is activated per-dataset based on a task-type flag.

This has a practical benefit: at fixed per-model recipe, training one unified model is roughly half the cost of training two specialized ones. The shared column embedding and row interaction stages also benefit from seeing both classification and regression data during pretraining, potentially improving feature representations. At inference, the model routes to the correct head based on the task specification.

\subsection{Single-Stage Pretraining}
\label{sec:singlestage}

Prior tabular foundation models rely on multi-stage curriculum learning to train stably. TabICL~v2 uses three stages: 500K steps on small datasets (1{,}024 samples), then 40K steps scaling to 10K samples, then 10K steps scaling to 60K samples. Each stage requires careful learning rate adjustment and consumes significant compute.

\tabh eliminates this curriculum entirely. From the first training step, the model sees datasets with at least 2{,}048 rows, 2--100 features, and a mix of classification and regression tasks. The configured maximum sequence length is 12{,}288 rows for \tabhv and increases to 14{,}336 for \tabhvone. This is possible because of four architectural stability mechanisms that prevent the gradient instabilities which necessitate curriculum staging in the first place:

\paragraph{Bounded QASSMax.} TabICL~v2 introduced QASSMax, which rescales attention queries by $\text{MLP}_{\text{base}}(\log n) \cdot (1 + \tanh(\text{MLP}_{\text{gate}}(q)))$. The unbounded MLP output can grow arbitrarily during training, eventually undoing normalization and causing instability at long sequence lengths. In \tabh, we apply a hard bound to the base scaling:
\begin{equation}
    \text{scale}_{\text{base}} = c_{\max} \cdot \tanh\!\left(\frac{\text{MLP}_{\text{base}}(\log n)}{c_{\max}}\right)
\end{equation}
where $c_{\max} = 10$ by default. This guarantees that scaling factors remain in $[-c_{\max}, c_{\max}]$ regardless of parameter magnitude, preserving the $\log n$ awareness and query-dependent gating of QASSMax. At initialization, $\tanh(x/c_{\max}) \approx x/c_{\max}$ so behavior is identical to the unbounded variant; the bound only activates when parameters grow large.

\paragraph{Inter-stage normalization.} RMSNorm~\cite{zhang2019rmsnorm} layers are inserted between the three processing stages, constraining activation magnitudes and preventing residual stream growth from propagating across stage boundaries.

\paragraph{Learnable residual scaling.} We rescale the residual stream before each block and re-inject the stack input via a small learnable weight. Within each of the three transformer stacks (column embedding, row interaction, ICL), let $x_0$ denote that stack's input and $x_{i-1}$ the residual stream entering block $i$. The block input is
\begin{equation}
    \tilde{x}_i = \lambda_{\text{resid}}^{(i)} \cdot x_{i-1} + \lambda_{x_0}^{(i)} \cdot x_0,
    \qquad
    x_i = \text{block}_i(\tilde{x}_i),
\end{equation}
where $\text{block}_i(\cdot)$ retains its own internal (pre-norm) residual connection. The per-block scalars are learnable, initialized linearly with depth as $\lambda_{\text{resid}}^{(i)}\!: 1.0\!\to\!0.9$ and $\lambda_{x_0}^{(i)}\!: 0.1\!\to\!0.02$, and are independent across the three stages. This gives the optimizer a direct knob on residual magnitude in deep stacks while preserving a learnable shortcut from each stack's input to its deeper layers.

\paragraph{Logit soft-capping.} Classification logits are passed through $c \cdot \tanh(\text{logits}/c)$ with $c = 30$, preventing extreme logit values that destabilize softmax cross-entropy gradients during early training when the model sees highly variable data.

\medskip
\noindent Together, these mechanisms make training stable enough that \tabhv reaches competitive performance in 100K optimizer updates ($\sim$6.4M synthetic datasets) without any curriculum scheduling, compared to TabICL~v2's 550K steps ($\sim$35M datasets) across three carefully tuned stages. As a rough proxy for pretraining cost, this is roughly $5.5\times$ fewer synthetic datasets per model; in addition, our unified head removes the need to train a second model, contributing a further $\sim$2$\times$ factor when the goal is to ship both classification and regression. We caution that ``synthetic datasets seen'' is an imperfect proxy for FLOPs or wall-clock time: differences in batch size, sequence length, model size, and pipeline efficiency can shift the true cost ratio in either direction, and a direct compute comparison would require pipeline-level data not currently public for TabICL~v2.

\subsection{Noise-Aware Pretraining}

In real-world datasets, not every feature is informative. To teach the model robustness to irrelevant features, \tabh's DAG-SCM generator can replace output dimensions with independent noise sampled from the same distribution as root causes. The configured maximum is 3 dimensions per node for \tabhv and 2 for \tabhvone. This creates features that are statistically plausible but carry no signal. TabICL~v2 does not explicitly inject noise dimensions in this manner.

\section{Pretraining}
\label{sec:pretraining}

\tabh is pretrained on synthetically generated tabular datasets produced by a DAG-based structural causal model (DAG-SCM)~\cite{pearl2009causality}. We use the same prior framework as TabICL~v2~\cite{qu2026tabiclv2}: random directed acyclic graphs where each node applies a randomly sampled transformation to its parent nodes.

\begin{figure}[t]
  \centering
  \includegraphics[width=\columnwidth]{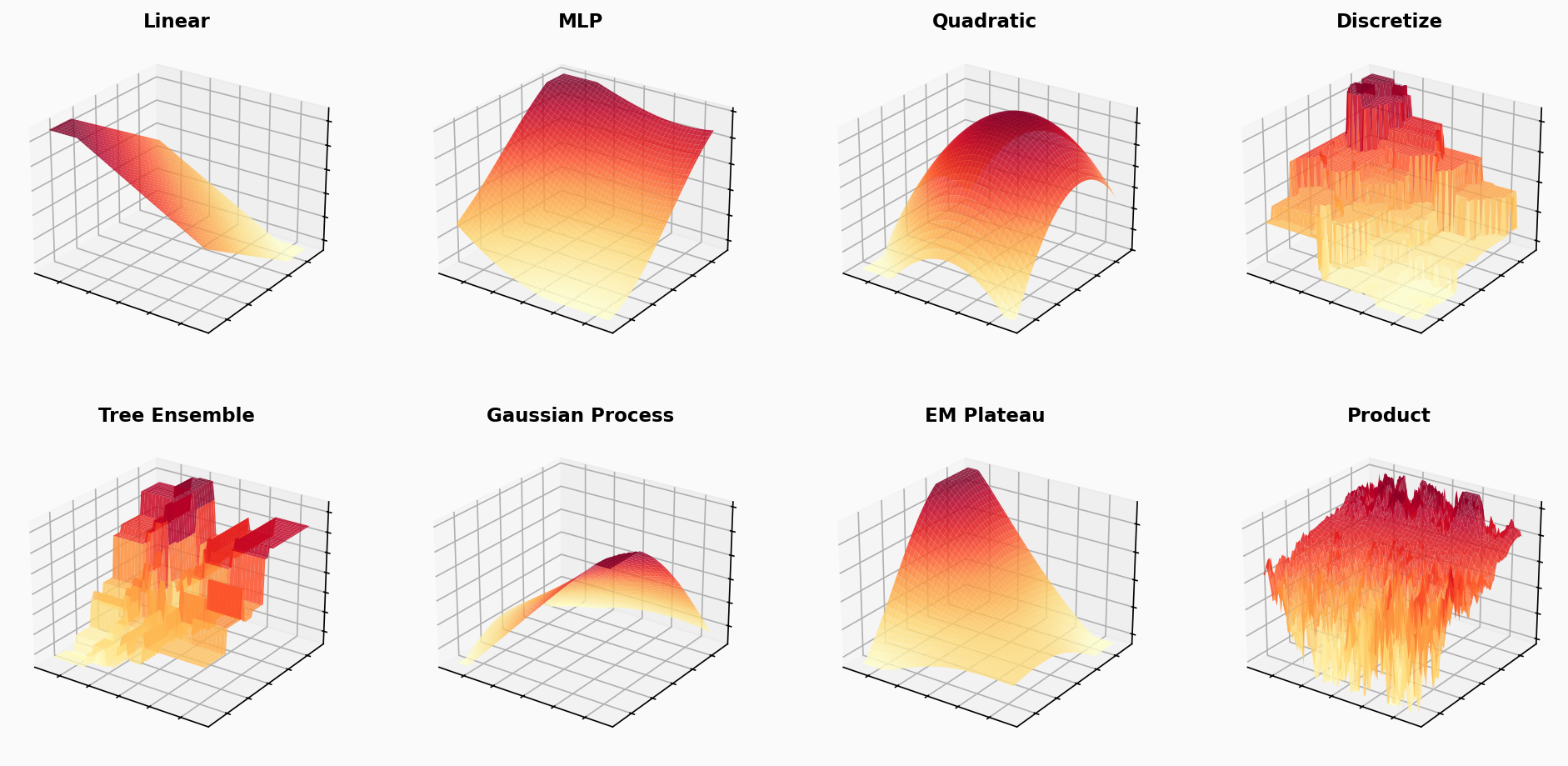}
  \caption{\textbf{Random function types in DAG-SCM.} Each of the 8 function types produces distinct surface shapes (2D input $\to$ 1D output). These are the same function types used in TabICL~v2~\cite{qu2026tabiclv2}.}
  \label{fig:functions}
\end{figure}

\paragraph{Synthetic data generation.} Each dataset is generated by: (1)~sampling a random DAG structure with configurable edge density and maximum of 10 parents per node; (2)~assigning each node one of 8 function types (linear, MLP, quadratic, discretize, tree ensemble, Gaussian process, EM plateau, product); (3)~sampling root values from diverse distributions; (4)~propagating through the DAG. Figure~\ref{fig:functions} visualizes representative surfaces for each function type.

Post-processing includes z-score standardization (zero mean, unit variance, clipped to $[-100, 100]$), optional regression-to-classification conversion, and an ExtraTrees~\cite{geurts2006extratrees}-based quality filter that rejects datasets where a simple model cannot beat a constant baseline (filtering ${\sim}$35\% of classification and ${\sim}$25\% of regression datasets). An additional filter rejects datasets where ExtraTrees achieves AUC $>0.99$, as overly easy datasets provide minimal learning signal.

\paragraph{Training configuration.} As described in Section~\ref{sec:singlestage}, both checkpoints use one training stage with batch size 64 (micro-batch 4 across 8 GPUs), 2--100 features per dataset, and uniformly sampled sequence lengths. \tabhv uses 100K optimizer updates ($\sim$6.4M synthetic datasets), configured sequence-length bounds of 2{,}048 and 12{,}288 rows, and learning rate $8 \times 10^{-4}$. \tabhvone uses 50K optimizer updates with four-step gradient accumulation---equivalent to 200K non-accumulated batch steps and $\sim$12.8M synthetic datasets---and increases the configured maximum sequence length to 14{,}336 rows while using learning rate $4 \times 10^{-4}$. Both use the Muon optimizer~\cite{jordan2024muon}, weight decay 0.05, gradient clipping at 10.0, cosine warmup scheduling (2\% warmup), bfloat16 mixed precision, and gradient checkpointing. No curriculum staging is required.

\section{Inference}
\label{sec:inference}

\paragraph{Preprocessing.} \tabh applies automatic preprocessing: ordinal encoding for categorical columns, mean imputation for missing values, z-score normalization with clipping to $[-100, 100]$ (matching the training distribution), and outlier removal with a configurable z-score threshold (default 4.0).

\paragraph{Ensemble inference.} To improve robustness, \tabh generates $N$ ensemble members (default $N{=}8$) by varying: (1)~feature permutations via Latin square patterns (each ensemble member uses a row of a Latin square over the feature indices, so that across the $N$ members each feature appears in $N$ distinct input positions), (2)~target flipping on alternating members, and (3)~random feature sign flips. For classification, logits are averaged before softmax; for regression, quantile predictions are averaged.

\paragraph{Memory-efficient processing.} For large datasets, \tabh supports chunked processing: training data is split into subsets, predictions are computed per-chunk, and results are averaged. Combined with KV caching and activation offloading, this enables datasets with up to 500{,}000 rows on a single GPU.

\section{Evaluation}
\label{sec:evaluation}

\subsection{Benchmark Setup}

We evaluate \tabh on \textbf{TALENT}~\cite{liu2024talent}, spanning 300 classification and regression datasets. We report two checkpoints from the same unified model family: \tabhv (100K optimizer updates) and \tabhvone (50K optimizer updates with four-step gradient accumulation, equivalent to 200K non-accumulated batch steps). \tabhvone uses a larger synthetic-data budget and increases the configured maximum sequence length from 12{,}288 to 14{,}336 rows (Section~\ref{sec:pretraining}). We chose TALENT over TabArena~\cite{erickson2025tabarena} as the primary large-scale benchmark for three reasons: (1)~\emph{scale}---300 datasets vs.\ 51 provides far more statistical power; (2)~\emph{robustness}---no single difficult dataset can disproportionately affect rankings; (3)~\emph{diversity}---a mix of classification and regression tasks across diverse domains. We provide more details about the TALENT benchmark in the Appendix~\ref{sec:talentbench}.
We additionally report TabArena for \tabhvone (Appendix~\ref{sec:tabarenabench}; Section~\ref{sec:tabh2o_v11}).

Models are ranked per task; lower average rank is better. We compare against tuned CatBoost~\cite{prokhorenkova2018catboost}, LightGBM~\cite{ke2017lightgbm}, H2O AutoML~\cite{h2o2024automl}, TabPFN~v2.6~\cite{hollmann2025tabpfnv2}, and TabICL~v2~\cite{qu2026tabiclv2}.
To avoid related checkpoints affecting each other's ranks, we evaluate \tabhv and \tabhvone separately against the same baselines; their independently computed ranks are marked with an asterisk in Figures~\ref{fig:rank} and~\ref{fig:rank_tabarena_v11}.

\subsection{Main Results}

\definecolor{barfm}{HTML}{9E9E9E}
\definecolor{barth2o}{HTML}{F5C518}
\definecolor{barml}{HTML}{CCCCCC}

\begin{figure}[t]
  \centering
  \begin{tikzpicture}[yscale=0.72]
    \foreach \x in {0,1,...,6} {
      \draw[gray!20, thin] (\x,0.3) -- (\x,6.7);
    }
    \newcommand{\bh}{0.22}
    \fill[barml, rounded corners=1.5pt] (0,1-\bh) rectangle (4.98,1+\bh);
    \node[font=\scriptsize\bfseries, anchor=west] at (4.98+0.08,1) {4.98};
    \node[font=\small, anchor=east] at (-0.15,1) {LightGBM};
    \fill[barml, rounded corners=1.5pt] (0,2-\bh) rectangle (4.28,2+\bh);
    \node[font=\scriptsize\bfseries, anchor=west] at (4.28+0.08,2) {4.28};
    \node[font=\small, anchor=east] at (-0.15,2) {H2O AutoML};
    \fill[barml, rounded corners=1.5pt] (0,3-\bh) rectangle (4.00,3+\bh);
    \node[font=\scriptsize\bfseries, anchor=west] at (4.00+0.08,3) {4.00};
    \node[font=\small, anchor=east] at (-0.15,3) {CatBoost (tuned)};
    \fill[barfm, rounded corners=1.5pt] (0,4-\bh) rectangle (2.80,4+\bh);
    \node[font=\scriptsize\bfseries, anchor=west] at (2.80+0.08,4) {2.80};
    \node[font=\small, anchor=east] at (-0.15,4) {TabPFN v2.6};
    \fill[barth2o, rounded corners=1.5pt] (0,5-\bh) rectangle (2.55,5+\bh);
    \node[font=\scriptsize\bfseries, anchor=west] at (2.55+0.08,5) {2.55$^\ast$};
    \node[font=\small, anchor=east] at (-0.15,5) {TabH2O v1};
    \fill[barth2o, rounded corners=1.5pt] (0,6-\bh) rectangle (2.54,6+\bh);
    \node[font=\scriptsize\bfseries, anchor=west] at (2.54+0.08,6) {2.54$^\ast$};
    \node[font=\small\bfseries, anchor=east] at (-0.15,6) {TabH2O v1.1};
    \fill[barfm, rounded corners=1.5pt] (0,7-\bh) rectangle (2.13,7+\bh);
    \node[font=\scriptsize\bfseries, anchor=west] at (2.13+0.08,7) {2.13};
    \node[font=\small, anchor=east] at (-0.15,7) {TabICL v2};
    \draw[gray!40] (0,0.5) -- (6.3,0.5);
    \foreach \x in {0,1,...,6} {
      \draw[gray!40] (\x,0.5) -- (\x,0.38);
      \node[font=\small, gray!70!black] at (\x,0.1) {\x};
    }
    \node[font=\small, gray!60!black] at (3,-0.35) {Average Rank (lower is better)};
    \fill[barml, rounded corners=1pt] (3.8,7.55) rectangle (4.15,7.75);
    \node[font=\scriptsize, anchor=west] at (4.22,7.65) {Traditional ML};
    \fill[barfm, rounded corners=1pt] (3.8,7.2) rectangle (4.15,7.4);
    \node[font=\scriptsize, anchor=west] at (4.22,7.3) {Foundation Model};
    \fill[barth2o, rounded corners=1pt] (3.8,6.85) rectangle (4.15,7.05);
    \node[font=\scriptsize, anchor=west] at (4.22,6.95) {TabH2O};
  \end{tikzpicture}
  \caption{\textbf{Average rank on TALENT} (300 datasets, 3 task types, 6 methods per comparison, lower rank is better). \tabhvone outperforms every traditional ML approach with zero hyperparameter tuning. TabICL~v2 ranks highest among all models. TabICL~v2 uses separate classification and regression models (two full training runs), while \tabhvone uses a single unified model. \textsuperscript{*}\tabhv and \tabhvone are each ranked independently against the same five baselines; baseline bars shown correspond to the \tabhvone comparison.}
  \label{fig:rank}
\end{figure}
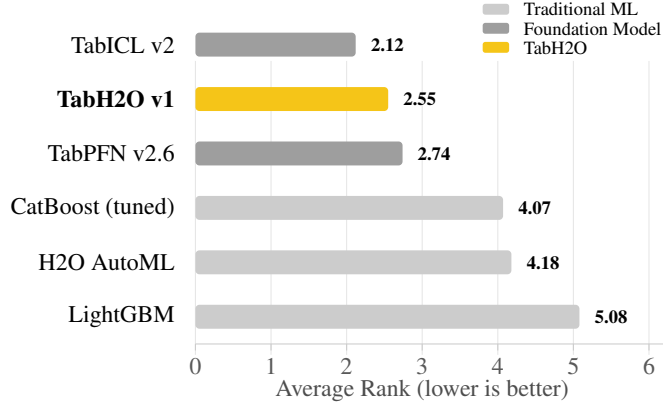

\begin{figure}[t]
  \centering
  \includegraphics[width=0.7\linewidth]{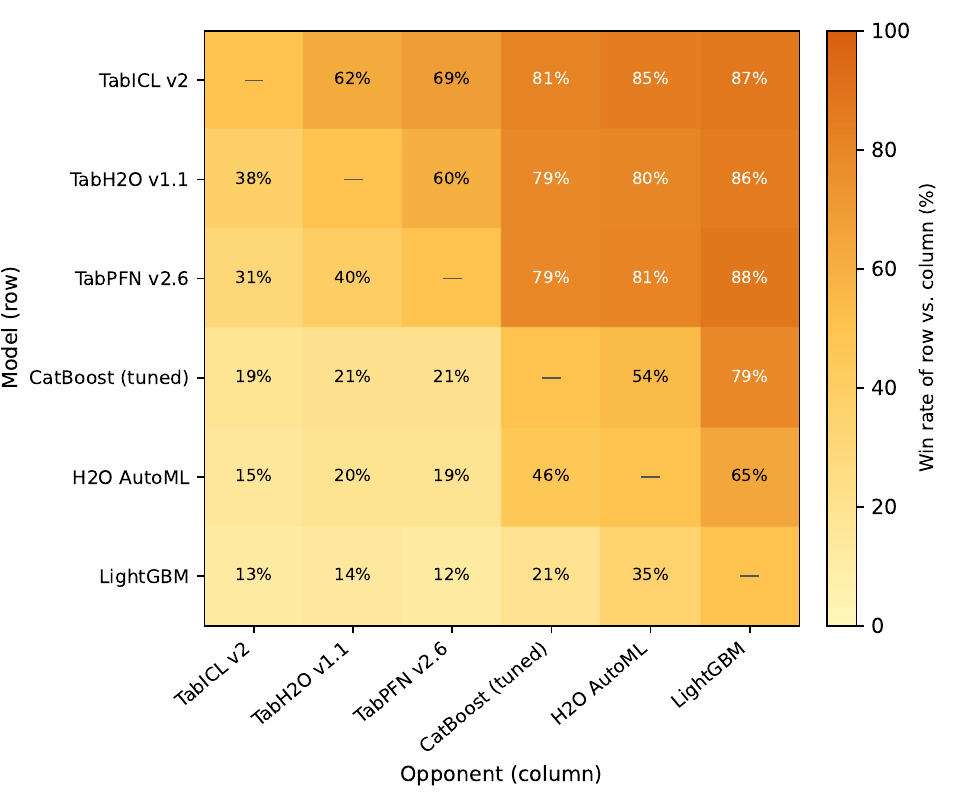}
  \caption{\textbf{Pairwise Winrates} on the TALENT Benchmark (\tabhvone). Note that TabICL~v2 uses two models, while \tabhvone is a single unified model.}
  \label{fig:winmatrix}
\end{figure}

Figure~\ref{fig:rank} presents results on the full TALENT benchmark. \tabhvone achieves an average rank of 2.54, placing in the top~3 on 80\% of datasets. The key findings:

\begin{itemize}[nosep]
    \item \tabhvone outperforms every traditional ML approach---tuned CatBoost (4.00), H2O AutoML (4.28), LightGBM (4.98)---with \emph{zero hyperparameter tuning}.
    \item \tabhvone is competitive with TabPFN~v2.6 (rank 2.80) while being a single unified model.
    \item TabICL~v2 achieves the best average rank (2.13) with separate specialized models and substantially more pretraining compute (Sec~\ref{sec:singlestage}).
    \item As a unified model trained with substantially less compute, \tabhvone's rank of 2.54 demonstrates the viability of the unified approach---see the win-rate matrix in Figure~\ref{fig:winmatrix}.
\end{itemize}

\subsection{TabArena Results}
\label{sec:tabh2o_v11}

On TabArena~\cite{erickson2025tabarena} (51 datasets; Appendix~\ref{sec:tabarenabench}), we compare \tabhvone against TabICL~v2 and TabPFN~v2.6 under the same average-rank protocol. Restricting TabArena to foundation models keeps the comparison fair and avoids mixing unequal AutoML time budgets. \tabhvone achieves the best average rank of 1.94, ahead of TabPFN~v2.6 (2.00) and TabICL~v2 (2.06). Figure~\ref{fig:rank_tabarena_v11} shows the ranking.

In its separate comparison, \tabhv ranks 2.12, behind TabPFN~v2.6 (1.90) and TabICL~v2 (1.98), while \tabhvone narrowly leads both models. TabPFN~v2.6 has a stronger relative standing on TabArena than on TALENT, highlighting benchmark-dependent performance and the value of reporting both benchmarks.

\begin{figure}[t]
  \centering
  \begin{tikzpicture}[yscale=0.72]
    \foreach \x in {0,1,...,3} {
      \draw[gray!20, thin] (\x,0.3) -- (\x,3.7);
    }
    \newcommand{\bh}{0.22}
    \fill[barth2o, rounded corners=1.5pt] (0,1-\bh) rectangle (2.12,1+\bh);
    \node[font=\scriptsize\bfseries, anchor=west] at (2.12+0.08,1) {2.12$^\ast$};
    \node[font=\small, anchor=east] at (-0.15,1) {TabH2O v1};
    \fill[barfm, rounded corners=1.5pt] (0,2-\bh) rectangle (2.06,2+\bh);
    \node[font=\scriptsize\bfseries, anchor=west] at (2.06+0.08,2) {2.06};
    \node[font=\small, anchor=east] at (-0.15,2) {TabICL v2};
    \fill[barfm, rounded corners=1.5pt] (0,3-\bh) rectangle (2.00,3+\bh);
    \node[font=\scriptsize\bfseries, anchor=west] at (2.00+0.08,3) {2.00};
    \node[font=\small, anchor=east] at (-0.15,3) {TabPFN v2.6};
    \fill[barth2o, rounded corners=1.5pt] (0,4-\bh) rectangle (1.94,4+\bh);
    \node[font=\scriptsize\bfseries, anchor=west] at (1.94+0.08,4) {1.94$^\ast$};
    \node[font=\small\bfseries, anchor=east] at (-0.15,4) {TabH2O v1.1};
    \draw[gray!40] (0,0.5) -- (3.5,0.5);
    \foreach \x in {0,1,...,3} {
      \draw[gray!40] (\x,0.5) -- (\x,0.38);
      \node[font=\small, gray!70!black] at (\x,0.1) {\x};
    }
    \node[font=\small, gray!60!black] at (1.75,-0.35) {Average Rank (lower is better)};
  \end{tikzpicture}
  \caption{\textbf{Average rank on TabArena} (51 datasets, 3 foundation models per comparison, lower is better). Among TabPFN~v2.6, TabICL~v2, and \tabh, \tabhvone ranks first. \textsuperscript{*}\tabhv and \tabhvone are each ranked independently against the same two baselines; baseline bars shown correspond to the \tabhvone comparison.}
  \label{fig:rank_tabarena_v11}
\end{figure}

\subsection{Inference Speed}

\tabh has no training step---the entire inference operation is a single forward pass through a pretrained transformer. We benchmark end-to-end latency (including network round-trip) and server-side GPU~\footnote{We report inference metrics using an NVIDIA RTX PRO 6000.} inference time across a grid of dataset sizes (10--500K rows, 2--100 features), with a 50/50 train/test split.

\begin{figure}[t]
  \centering
  \includegraphics[width=\columnwidth]{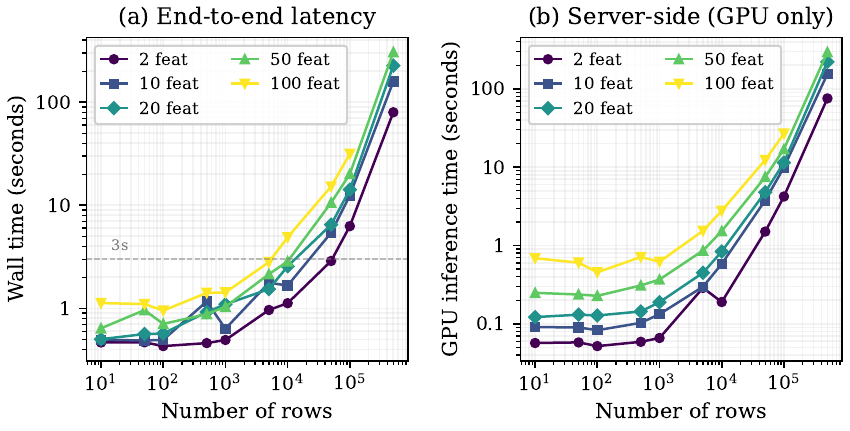}
  \caption{\textbf{Prediction latency vs.\ dataset size.} (a)~End-to-end wall time including network round-trip to the hosted API. The dashed line marks 3 seconds. (b)~Server-side GPU inference time only. For datasets up to 10K rows, end-to-end latency is dominated by network overhead ($\sim$0.4s baseline); GPU inference itself completes in under 1 second.}
  \label{fig:latency}
\end{figure}

Figure~\ref{fig:latency} shows latency scaling. Key observations:
\begin{itemize}[nosep]
    \item \textbf{Sub-3-second predictions} for datasets up to $\sim$10K rows regardless of feature count---covering most real-world use cases (customer churn, fraud signals, clinical data).
    \item \textbf{Latency scales with $n^2$} (quadratic in rows) in the unchunked regime, as expected from the ICL attention complexity, while feature count has a smaller effect ($nm^2$ term).
    \item \textbf{Network overhead dominates for small datasets}: the $\sim$0.4s baseline is API round-trip; GPU inference for 1K rows $\times$ 20 features takes only 189ms.
    \item \textbf{500K rows remain feasible}: the largest configuration (500K $\times$ 50) completes in $\sim$5 minutes with chunked processing.
\end{itemize}

Table~\ref{tab:speed} shows representative end-to-end times for common dataset sizes.

\begin{table}[h]
    \centering
    \caption{\textbf{Representative prediction latency.} End-to-end wall-clock time (including network) and server-side GPU time for selected dataset configurations.}
    \label{tab:speed}
    \small
    \begin{tabular}{rrrcr}
        \toprule
        \textbf{Rows} & \textbf{Features} & \textbf{Total Cells} & \textbf{Wall Time} & \textbf{GPU Time} \\
        \midrule
        1{,}000 & 20  & 21K     & 1.1\,s    & 0.19\,s \\
        5{,}000 & 20  & 105K    & 1.5\,s    & 0.45\,s \\
        10{,}000 & 20 & 210K    & 2.6\,s    & 0.84\,s \\
        10{,}000 & 50 & 510K    & 2.9\,s    & 1.5\,s \\
        50{,}000 & 20 & 1.05M   & 6.5\,s    & 4.8\,s \\
        50{,}000 & 100 & 5.05M  & 15.2\,s   & 12.4\,s \\
        100{,}000 & 20 & 2.1M   & 14.2\,s   & 11.4\,s \\
        100{,}000 & 100 & 10.1M & 31.5\,s   & 26.8\,s \\
        500{,}000 & 20 & 10.5M  & 3.8\,min  & 3.7\,min \\
        \bottomrule
    \end{tabular}
\end{table}

\begin{figure}[t]
  \centering
  \includegraphics[width=0.85\columnwidth]{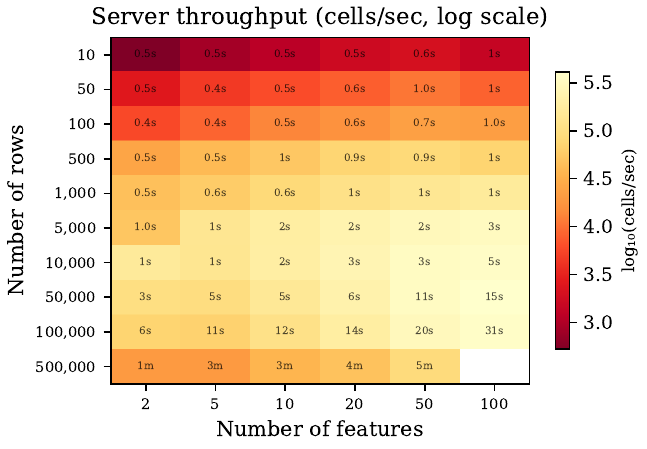}
  \caption{\textbf{Wall time heatmap} across all tested configurations (rows $\times$ features). Cell annotations show end-to-end time. Color encodes server throughput (cells/sec) on a log scale---lighter is faster.}
  \label{fig:heatmap}
\end{figure}

Figure~\ref{fig:heatmap} provides a complete overview of wall times across the full grid of configurations.

\section{Task Support}
\label{sec:tasks}

The unified architecture enables \tabh to support multiple task types natively:

\begin{enumerate}
    \item \textbf{Classification}: Binary and multi-class (up to 10 natively; hierarchical decomposition for more). Returns predictions and per-class probabilities.
    \item \textbf{Regression}: Predicts 999 quantiles, providing point estimates (mean, median), variance, and calibrated confidence intervals at arbitrary levels.
    \item \textbf{Time series forecasting}: Historical data with a time column is transformed into supervised features; the model produces future predictions with confidence intervals.
    \item \textbf{Anomaly detection}: Rows are scored by their fit to the data distribution via the predicted quantile density, without requiring anomaly labels.
    \item \textbf{Clustering \& embeddings}: Fixed-size row embeddings extracted from the row interaction stage can be used for unsupervised grouping.
    \item \textbf{Missing value imputation}: Missing values are predicted conditioned on observed features by treating each missing column as a regression target.
\end{enumerate}

\section{Conclusions}

We present \tabh, a unified foundation model for tabular prediction built on the TabICL architecture family. By training a single model for both classification and regression, introducing training stability mechanisms, and incorporating noise-aware pretraining, \tabhv achieves strong performance on the 300-task TALENT benchmark while using substantially less pretraining compute than TabICL~v2 across two specialized models.
\tabhvone preserves the unified approach with a larger single-stage pretraining budget ($\sim$12.8M synthetic datasets) and a higher configured maximum sequence length, remaining competitive on TALENT while attaining the best average rank among the compared methods on TabArena.

Our work is part of the growing ecosystem of tabular foundation models, and we gratefully acknowledge the foundational contributions of TabPFN~\cite{hollmann2023tabpfn, hollmann2025tabpfnv2} and TabICL~\cite{qu2025tabicl, qu2026tabiclv2}. We believe that continued iteration on training stability, data diversity, and unified multi-task learning can close the remaining gap to specialized models while maintaining the practical advantages of zero-configuration prediction.

\tabh is freely available at \url{https://tabh2o.h2oai.com}.

\section{The TabH2O API}
\label{sec:api}

\tabh is available as a hosted API with a free tier:

\begin{lstlisting}[language=bash]
curl -X POST https://tabh2o.h2oai.com/api/v1/predict \
  -H "Authorization: Bearer $API_KEY" \
  -H "Content-Type: application/json" \
  -d '{
    "task": "classification",
    "target_column": "purchased",
    "train": {
      "columns": ["age", "income", "purchased"],
      "data": [
        [25, 50000, "Yes"], [30, 60000, "No"],
        [45, 80000, "Yes"], [22, 35000, "No"]
      ]
    },
    "test": {
      "columns": ["age", "income"],
      "data": [[27, 52000], [38, 71000]]
    }
  }'
\end{lstlisting}

{
\small
\bibliographystyle{plain}
\bibliography{references}
}


\appendix
\section{TALENT Benchmark Details}
\label{sec:talentbench}

The TALENT benchmark~\citep{liu2024talent} consists of 300 tabular datasets covering three types of tasks: (i) \textbf{Binary classification} with 120 datasets, (ii) \textbf{Multiclass classification} with 80 datasets, and (iii) \textbf{Regression} with 100 datasets. 

For each dataset, the data is split into training, validation, and test sets with ratios of 64\%, 16\%, and 20\%, respectively. The final results of \tabh are reported on the held-out test split.

\paragraph{Evaluation metrics.}
We follow the evaluation protocol defined by TALENT~\citep{liu2024talent}. Accuracy is used as the main metric for classification problems, while root mean squared error (RMSE) is used for regression. 

In particular, we use the \textbf{Average rank} as the primary summary metric. For each dataset, methods are ordered based on accuracy or RMSE depending on the task, where rank 1 corresponds to the best-performing model.
The average rank of a method $m$ is computed as the mean of its ranks over all datasets $\mathcal{D}$:
\begin{equation}
\text{AvgRank}(m) = \frac{1}{|\mathcal{D}|} \sum_{i \in \mathcal{D}} \text{rank}_i(m) 
\end{equation}

Thus, a lower average rank indicates stronger overall performance across datasets.

\paragraph{Methods compared.}
We compare \tabhvone against five strong baselines covering the three dominant approaches to tabular prediction---gradient-boosted trees, AutoML systems, and tabular foundation models:
\begin{itemize}[nosep]
    \item \textbf{LightGBM (default)}~\citep{ke2017lightgbm}: gradient-boosted decision trees with default hyperparameters.
    \item \textbf{CatBoost (tuned)}~\citep{prokhorenkova2018catboost}: gradient-boosted decision trees with per-dataset hyperparameter tuning.
    \item \textbf{H2O AutoML}~\citep{h2o2024automl}: full AutoML pipeline (model selection, ensembling, stacking) with a 10-minute training budget per dataset.
    \item \textbf{TabPFN v2.6}~\citep{hollmann2025tabpfnv2}: the latest publicly released TabPFN package version at evaluation time, used with default inference settings.
    \item \textbf{TabICL v2}~\citep{qu2026tabiclv2}: the publicly released TabICL v2 weights (separate classification and regression checkpoints), evaluated with default inference settings.
\end{itemize}
All foundation models (\tabhvone, TabPFN v2.6, TabICL v2) are evaluated in zero-shot mode---no per-dataset hyperparameter tuning or fine-tuning is performed.

\paragraph{Note on rank values.}
Average rank values depend on the set of methods compared. We evaluate \tabhv and \tabhvone separately against the same five baselines, yielding two six-method comparisons with ranks in $[1, 6]$. This prevents checkpoints from the same model family from taking ranks from one another. In figures containing both checkpoints, checkpoint ranks are marked with an asterisk and baseline ranks correspond to the \tabhvone comparison. The average ranks reported here are therefore not directly comparable with those reported by TabICL~v2~\cite{qu2026tabiclv2} on TALENT, which include a different set of baselines. Within each common set of methods, lower rank still indicates better predictive performance.
\section{TabArena Benchmark Details}
\label{sec:tabarenabench}

The TabArena benchmark~\citep{erickson2025tabarena} consists of 51 tabular datasets: 38 classification datasets with at most 10 classes and 13 regression datasets. Evaluation follows the TabArena protocol of repeated cross-validation, with ROC~AUC for binary classification, log-loss for multiclass classification, and RMSE for regression.

\paragraph{Evaluation metrics.}
As on TALENT (Appendix~\ref{sec:talentbench}), we report \textbf{Average rank} as the primary summary metric. For each dataset, methods are ordered by the official TabArena metric for that task (higher-is-better for AUC; lower-is-better for log-loss and RMSE), and rank~1 denotes the best method. The average rank of a method is the mean of its per-dataset ranks over the 51 tasks. With the three foundation models compared below, ranks lie in $[1, 3]$.

\paragraph{Methods compared.}
On TabArena we compare \tabhvone against the following foundation-model baselines under the same zero-shot / default settings used elsewhere in this report:
\begin{itemize}[nosep]
    \item \textbf{TabPFN v2.6}~\citep{hollmann2025tabpfnv2}: publicly released TabPFN package version, default inference settings.
    \item \textbf{TabICL v2}~\citep{qu2026tabiclv2}: publicly released TabICL~v2 weights (separate classification and regression checkpoints), default inference settings.
\end{itemize}
All methods (\tabhvone, TabPFN v2.6, TabICL v2) are evaluated in zero-shot mode---no per-dataset hyperparameter tuning or fine-tuning is performed. All 51 tasks completed successfully for every method in this comparison. We evaluate \tabhv and \tabhvone separately against the same two baselines, yielding two three-method comparisons. This prevents checkpoints from the same model family from taking ranks from one another. In Figure~\ref{fig:rank_tabarena_v11}, checkpoint ranks are marked with an asterisk and baseline ranks correspond to the \tabhvone comparison.

\paragraph{Relation to TALENT.}
TabArena is smaller than TALENT (51 vs.\ 300 datasets) but uses repeated cross-validation and is a complementary public leaderboard for tabular foundation models. We therefore report TabArena alongside TALENT for \tabhvone (Section~\ref{sec:tabh2o_v11}), while keeping TALENT as the primary large-scale comparison. Traditional ML and AutoML baselines, including H2O AutoML with a 10-minute budget, are reported on TALENT (Appendix~\ref{sec:talentbench}).

\end{document}